\documentclass[letterpaper, 10 pt, journal, twoside]{IEEEtran}
\ifCLASSINFOpdf
\else
\fi

\usepackage{graphics} 
\usepackage{epsfig} 
\usepackage{mathptmx} 
\usepackage{times} 
\usepackage{amsmath} 
\usepackage{amssymb}  
\usepackage{siunitx}
\usepackage{epsfig}
\usepackage{bbding}
\usepackage{pifont}
\usepackage{wasysym}
\usepackage{amssymb}
\usepackage[font=small]{caption}
\usepackage{xcolor}
\newcommand*\colourcheck[1]{%
  \expandafter\newcommand\csname #1check\endcsname{\textcolor{#1}{\ding{52}}}%
}
\colourcheck{red}
\newcommand{\xmark}{\ding{55}}%
\DeclareMathOperator{\E}{\mathbb{E}}
\DeclareMathOperator*{\argmax}{argmax} 
\DeclareMathOperator*{\argmin}{argmin} 

\begin{document}
%
\title{
Deceiving Image-to-Image Translation Networks for Autonomous Driving with Adversarial Perturbations}
%
%
%

\author{Lin Wang, Wonjune Cho, and Kuk-Jin Yoon%
\thanks{Manuscript received: September 9, 2019; Revised December 5, 2019; Accepted January, 2, 2020. This paper was recommended for publication by Editor Cesar Cadena Lerma upon  evaluation of the Associate Editor and Reviewers' comments.  This work was supported by the National Research Foundation of Korea (NRF) grant funded by the Korea government (MSIT) (NRF-2018R1A2B3008640).} 
\thanks{The authors are with Visual Intelligence Lab., Department of Mechanical Engineering, KAIST, Korea (e-mail: {\footnotesize wanglin@kaist.ac.kr, wonjune@kaist.ac.kr, kjyoon@kaist.ac.kr}).}%
\thanks{Digital Object Identifier (DOI): see top of this page.}
}

\markboth{IEEE Robotics and Automation Letters. Preprint Version. Accepted January, 2020}
{Wang \MakeLowercase{\textit{et al.}}: Deceiving Image-to-Image Translation Networks for Autonomous Driving with Adversarial Perturbations} 

%



\maketitle

\begin{abstract}
Deep neural networks (DNNs) have achieved impressive performance on handling computer vision problems, however, it has been found that DNNs are vulnerable to adversarial examples. 
For such reason, adversarial perturbations have been recently studied in several respects. However, most previous works have focused on image classification tasks, and it has never been studied regarding adversarial perturbations on Image-to-image (Im2Im) translation tasks, 
 showing great success in handling paired and/or unpaired mapping problems in the field of autonomous driving and robotics. 
This paper examines different types of adversarial perturbations that can fool Im2Im frameworks for autonomous driving purpose. We propose both quasi-physical and digital adversarial perturbations that can make Im2Im models yield unexpected results. We then empirically analyze these perturbations and show that they generalize well under both paired for image synthesis and unpaired settings for style transfer. We also validate that there exist some perturbation thresholds over which the Im2Im mapping is disrupted or impossible. The existence of these perturbations reveals that there exist crucial weaknesses in Im2Im models. Lastly, we show that our methods illustrate how  perturbations affect the quality of outputs, pioneering the improvement of the robustness of current SOTA networks for autonomous driving.  
\end{abstract}
\begin{IEEEkeywords}
Im2Im, adversarial attack, autonomous driving.
\end{IEEEkeywords}

%
\IEEEpeerreviewmaketitle

\section{Introduction}
%
%
%
%
\IEEEPARstart{D}{eep} neural networks (DNNs) are remarkably successful on handling many computer vision tasks, however, they have been shown to be vulnerable to adversarial perturbations of inputs \cite{ kurakin2016adversarial}. 
The adversarial perturbations are physically and digitally distorted input examples that can attack and fool the learned model such that it would produce an intentionally fabricated or an unexpected result \cite{kurakin2016adversarial}. DNNs for many visual intelligence tasks, such as image classification \cite{kurakin2016adversarial, hendrik2017universal, moosavi2016deepfool}, segmentation \cite{moosavi2017universal}, and detection \cite{chenphysical}, are shown to be highly vulnerable to them.  \emph{However, it has never been examined how much and what kinds of perturbations are detrimental to image-to-image (Im2Im) tasks. Im2Im frameworks are essentially more complex and sophisticated than pure classification-related problems.} 
\begin{figure}[t!]
\begin{center}
\renewcommand{\tabcolsep}{1pt}
\begin{tabular}{@{}ccccc@{}}
    \includegraphics[width=21mm,height=12mm]{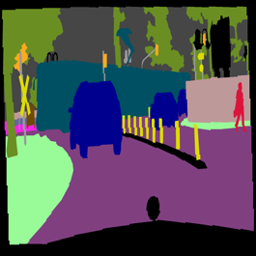}&
    \includegraphics[width=21mm,height=12mm]{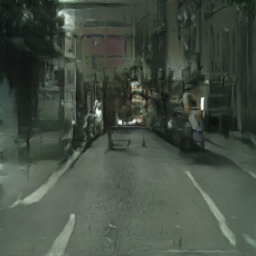}&
    \includegraphics[width=21mm,height=12mm]{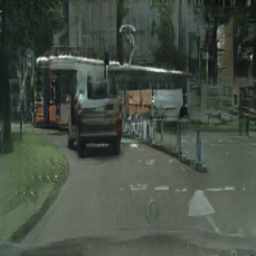}&
    \includegraphics[width=21mm,height=12mm]{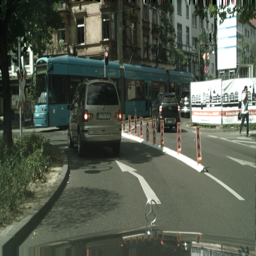}\\
    \includegraphics[width=21mm,height=12mm]{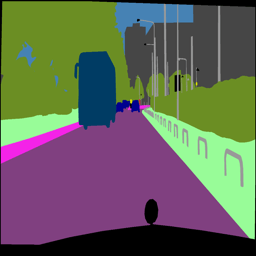}&
    \includegraphics[width=21mm,height=12mm]{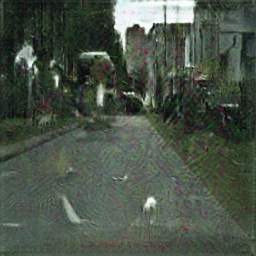}&
    \includegraphics[width=21mm,height=12mm]{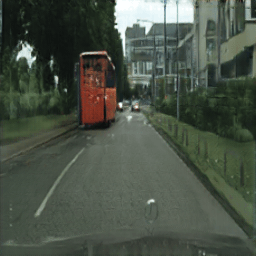}&
    \includegraphics[width=21mm,height=12mm]{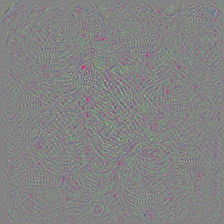}\\
    
    \includegraphics[width=21mm,height=12mm]{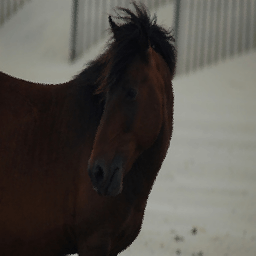}&
    \includegraphics[width=21mm,height=12mm]{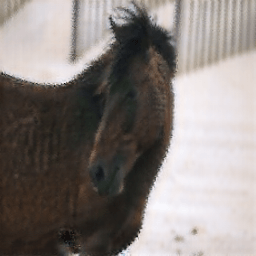}&
    \includegraphics[width=21mm,height=12mm]{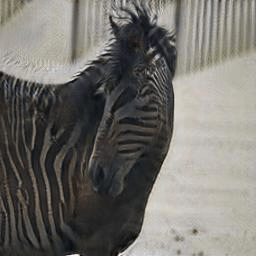}&
    \includegraphics[width=21mm,height=12mm]{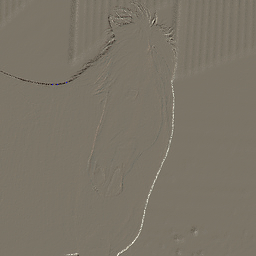}\\
\end{tabular}
\vspace{-5pt}
\caption{Examples of proposed adversarial perturbations. First row shows the result of a quasi-physical adversarial perturbation. From left to right, perturbed input ($R=2^\circ$), output with perturbation, output without perturbation, and target. Second row shows the result of an image-agnostic perturbation. From left to right, conditional input, output with perturbation, output without perturbation, and visualization of perturbation ($P_u=\{0.2,2000,\infty \}$). Third row shows the result of a flow-based perturbation on style transfer. From left to right, perturbed image, output with perturbation, output without perturbation, and visualization of perturbation ($\xi_{f}=1$). Our methods can fool Im2Im networks only with tiny perturbations. 
} 
\label{fig:impressive}
\end{center}
\vspace{-18pt}
\end{figure} 

Im2Im translation methods have shown considerable success in transforming images from one domain to another based on DNNs using paired  \cite{isola2017image,mostafavi2018event} or unpaired data \cite{zhu2017unpaired}. Recently, Im2Im networks have been broadly applied to the research of autonomous driving and robotics \cite{uricar2019yes,fabbrid}. For instance, Im2Im networks can be used for generating realistic environments in simulation for training self-driving car policies. Besides, Im2Im networks can be applied to study the way humans drive and either imitating human drivers exactly or inferring the goals that human drivers have and learning policies that accomplish the same goals better.  %
%

%
%

In this paper, we demonstrate the existence of physical and imperceptible or quasi-imperceptible digital perturbations leading to the malfunctioning of Im2Im tasks for autonomous driving applications. 
%
Actually, there exist many uncertainties in the current Im2Im frameworks regarding to their robustness and reliability. For instance, when a network fails to produce a desired output because of inconspicuous perturbations as in the \emph{horse2zebra} \cite{krizhevsky2012imagenet} examples shown in later sections, many would doubt the robustness of the network before questioning the dataset.
Therefore, it is quite imperative that we validate the reliability and robustness of Im2Im networks against different types of perturbations, especially to prepare the dataset and to design the networks for autonomous driving applications. To the best of our knowledge, this is the \textit{first} work about the adversarial perturbations to more complex Im2Im networks, focusing more on autonomous driving applications. 
We base our focus more on black-box attacks for Im2Im networks, where the adversary has no knowledge about the network structure as in \cite{papernot2016practical}, and propose both quasi-physical adversarial perturbations (i.e. using 2D transformation models) and digital adversarial perturbations (i.e. changing  pixel values or positions) that fool the Im2Im networks. \\
We summarize our main contributions as follows: 
(I) We are the \textit{first} to systematically present effective adversarial attack mechanisms for Im2Im, especially for autonomous driving. We show that there exist perturbations on images that can demolish the mapping from one domain to another, and through examining perturbations, this paper unlocks important factors that affect the mapping results but have not been considered before. 
(II) We propose \textit{three} novel 
  perturbation methods for both paired and unpaired settings, and show existing Im2Im networks are vulnerable to these perturbations.
%
(III) We find that quasi-physical adversarial perturbations actually matter seriously to the Im2Im networks. Meanwhile, we find there exist some perturbation \textit{patterns} and \textit{thresholds} of digital adversarial perturbations causing mapping to collapse. 
%
(IV) We 
\textit{empirically} analyze these perturbations and show that they generalize well under both paired and unpaired settings. We also show that our methods illustrate how the perceptual quality of outputs changes w.r.t the variation of perturbations. 

\section{Related works}
\noindent\textbf{Im2Im for autonomous driving and robotics.}
Im2Im aims at translating images from one domain to
another. In recent years, Im2Im frameworks have been applied to autonomous driving applications, such as semantic segmentation\cite{yang2018real}, scene understanding \cite{ghafoorian2018gan}, path planning\cite{uricar2019yes}, data augmentation \cite{soufi2019data}, etc. Meanwhile, they are also widely used for robots \cite{morrison2018closing,kuang2018using,rahmatizadeh2018vision} and medical imaging \cite{ma2019understanding, ma2017adversarial}. Im2Im networks are usually divided into two categories. 
The first one builds on the paired input-output images to learn a mapping model based on conditional generative adversarial networks (cGANs), as in \cite{isola2017image, mostafavi2018event}. 
%
The other one tackles the problem when there are no paired images, and instead, tries to learn a general mapping model in unpaired settings, as in  \cite{zhu2017unpaired,liu2017unsupervised} using unsupervised learning. 
The aforementioned approaches all rely on task-specific and predefined similarity functions between inputs and outputs. \emph{However, they never consider the reliability and robustness of the translation frameworks, which can be disrupted by the perturbations added to input and targeted images and are very crucial for autonomous driving and robotic applications.} 

\noindent\textbf{Adversarial perturbations.}
In general, perturbations can be categorized into two types: universal (image-agnostic) and image-dependent. \cite{moosavi2017universal,metzen2017universal} introduce universal perturbation methods based on DeepFool \cite{moosavi2016deepfool}.  
In contrast, \cite{poursaeed2018generative} present an image-dependent perturbation method for classification and detection tasks.
On the other hand, depending on the objectives of adversarial perturbations, they can be also divided into targeted and non-targeted perturbations. The goal of the targeted perturbation is to make the network produce a \emph{targeted output} with a perturbed image 
as in \cite{ kurakin2016adversarial,carlini2017towards}, while a non-targeted perturbation is to make the network produce \emph{any output} other than its original output (output without perturbation) as in \cite{papernot2016practical, moosavi2017universal,xiao2018spatially}. 
%
Furthermore, depending on the properties of perturbations, they can also be categorized into physical and digital domain perturbations \cite{eykholt2018robust}. The physical domain attacks aim to generate perturbations based on the geometric changes, as mentioned in \cite{athalye2017synthesizing,chenphysical}, which focus on real-world distortions due to different viewing distances and angles, lighting conditions, camera limitations\cite{chenphysical}, etc. On the other hand, digital attacks aim to change the image values with visually imperceptible modifications. 
\emph{In this paper, we consider both physical and digital domain perturbations for Im2Im since, in  reality, both perturbations exist and affect the security of autonomous driving and robotic applications}. 

\section{Background: adversarial attack for Im2Im}
\label{background_im2im}
In this work, we mainly focus on two representative Im2Im frameworks, the paired setting \cite{isola2017image} and the unpaired setting \cite{zhu2017unpaired},  commonly used for autonomous driving applications. 
An Im2Im model consists of a discriminator $D$ trained to classify images as real (from the dataset) or fake (generated), and a generator $G$ trained to fool the discriminator $D$. At inference time, the trained generator is used to generate fake images. 
The goal is to learn a mapping function from input domain $X$ (e.g. segmentation labels) to target domain $Y$ (e.g. color images) given samples $\{x_i\}_{i=1}^N$ where $x \in X$ and $\{y_i\}_{i=1}^M$ where $y \in Y$. For the paired mapping, $M=N$. We denote the data distribution as $x \sim p_{data}(x)$ and $y \sim p_{data}(y)$. Essentially, the fundamental objective functional for both paired and unpaired frameworks follows a minmax game as: 
\begin{equation}
\begin{split}
L_{Im2Im} = \max_{D}\min_{G}\E_{y \sim p_{data(y)}}[log(D(y)] + \\
       \E_{x \sim p_{data(x)}}[(1-log(D(G(x))]   
\end{split}
\end{equation}
Thus, when a generative model for Im2Im is trained, we can use $x \sim p_{data}(x)$ to generate images in domain $Y$.

Regarding adversarial perturbations for Im2Im, a naive way is to add perturbation to the $x \sim p_{data}(x)$ and then try to optimize the magnitude to get optimal perturbations. However, Im2Im is essentially different from general classification-based problems since it does not really have the classifier during the inference time. Besides, it is worth mentioning that there is a latent classifier, the discriminator which also determines the performance of Im2Im in training time. To this end, our approaches for attacking Im2Im frameworks are two folds. We first follow the generic way to apply adversarial perturbations to the generator $G$ only in inference time as shown in Fig.~\ref{fig:paired_univ_pert} (a),  Fig.~\ref{fig:spatial_pert} (a) and Fig.~\ref{fig:rotationImg}. We also propose to apply perturbations to the latent classifier. Although it takes the target domain images ($Y$) for discerning real from fake during training, it also needs to be considered since the adversarial perturbations may be maliciously added to the target images, and the mapping is usually bi-directional ($Y \to X$, $X \to Y$). Thus, as a new idea, we apply adversarial perturbations to the latent classifier in training time as shown in the bottom rows in Fig.~\ref{fig:paired_univ_pert} (b) and Fig.~\ref{fig:spatial_pert} (b). We aim to estimate how robust is the discriminator when the perturbations added to target images because these perturbations provide us guidance on how carefully should we prepare data and design Im2Im networks.  
\section{Proposed methods} 
Based on how Im2Im frameworks are applied in autonomous vehicle and robotics, we propose \emph{three} adversarial perturbation methods to fool the Im2Im networks by considering the geometric transformations of images \cite{eykholt2018robust, engstrom2017rotation}, modification of pixel values \cite{ moosavi2017universal, moosavi2016deepfool} and also spatial transformation \cite{xiao2018spatially, athalye2017synthesizing}.
We first focus on digital adversarial perturbations by adding imperceptible or quasi-imperceptible noise to the input images through optimization of perturbation. We then consider the quasi-physical adversarial perturbations 
such as transformations of images.

\subsection{Image-agnostic adversarial perturbation}

\begin{figure}[t!]
    \centering
    \includegraphics[width=7.5cm, height=4cm]{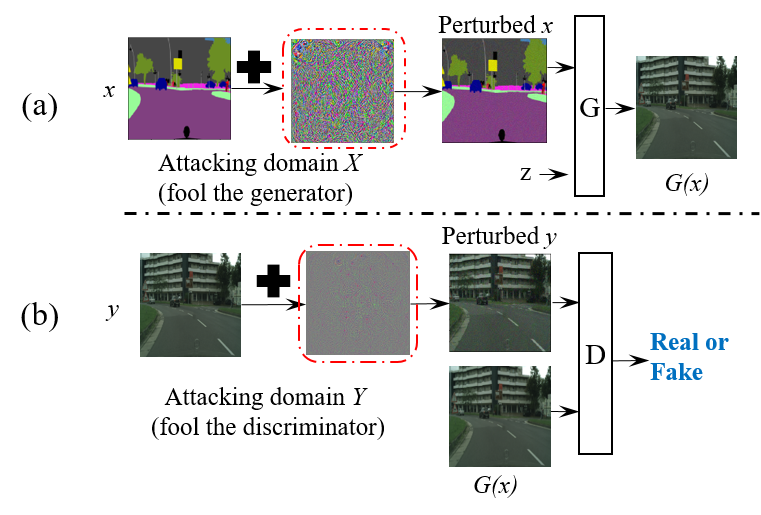}
    \vspace{-5pt}
    \caption{An illustration of image-agnostic perturbations for Im2Im (paired setting) via both domain $X$ (a) in inference time and domain $Y$ (b) in training time.  \textit{See more details in the contexts}.}
    \label{fig:paired_univ_pert}
    \vspace{-10pt}
\end{figure}

Since recent works, such as \cite{moosavi2017universal, moosavi2016deepfool,metzen2017universal}, have demonstrated that there exist image-agnostic adversarial perturbations that could cause the pretrained classifier to misclassify most of the perturbed inputs.
\textit{Here, we question whether an image-agnostic adversarial perturbation could cause any wrong mapping from one domain to another in Im2Im tasks, and if yes, to what extent does it affect domain translation? }
Hence, we focus on generating image-agnostic perturbations for images fed to Im2Im networks and investigate how they affect the mapping. 
\textit{Our aim is to ultimately fool the generator such that it fails to generate the expected outputs.} As mentioned in Sec.~\ref{background_im2im}, we first apply the image-agnostic adversarial perturbation to the generator in inference time as done in most previous works. We also aim to apply image-agnostic perturbation to the discriminator in the Im2Im models in training time, and then use pristine input domain images to test from the trained generator in inference time. 
Inspired by \cite{moosavi2017universal, hendrik2017universal}, which focus on the  classification problems, we propose algorithms for the challenging Im2Im problems and apply to autonomous driving. We consider the generation of image-agnostic adversarial perturbations under paired and unpaired settings. 
As shown in  Fig.~\ref{fig:paired_univ_pert} (a), the generator aims to map $x \sim p_{data}(x)$ (\textit{e.g.} segmentation labels) to $y \sim p_{data}(y)$ (\textit{e.g.} color images) under the paired setting. When image-agnostic perturbations are applied to the input domain $X$ in inference time, the generator takes the inputs and generate the color images in domain $Y$. 
When perturbations are added to $y \sim p_{data}(y)$, the discriminator learns to classify the synthesized and real \{$x$, perturbed $y$\} tuple. After training, $x \sim p_{data}(x)$ are then fed to generate color images in domain $Y$. 

We now formulate the image-agnostic perturbations and propose an approach on how to generate such perturbations for Im2Im. As shown in Fig.~\ref{fig:paired_univ_pert} (a), we first aim to fool the trained generator. 
Given a set of labels $X=\{x_1,..., x_n\}$ as inputs and color images $Y=\{x_1,..., x_n\}$ as targets in paired setting, where $n$ is the total number of images, we seek to generate perturbation $\Xi$ that can fool the  pretrained DNN-based classifier $C$ (e.g. VGG-19) on most images sampled from distribution $\mu$ in domain $X$ or $Y$ using the boundary threshold $\xi$, which can be formulated as $||\Xi||_p \le \xi$. Note that the labels predicted by $C$ do not need to be always true since we care more about finding the optimal perturbation for Im2Im tasks and our aim is then to find the minimal perturbation ${\Xi}_{min}$. Meanwhile, to guarantee that the constraint $||\Xi||_p \le \xi$ is always satisfied, we refer to the method proposed by \cite{moosavi2016deepfool, moosavi2017universal}, where the updated perturbation is further projected on the $l_p$ ball of radius $\xi$ and centered at zero. Denote $P_{\xi}(\Xi)$ as the projection operator defined as follows:
\vspace{-3pt}
\begin{equation}
P_{\xi}(\Xi)= \argmin_{\Xi'}||\Xi -\Xi'||_2  \quad \text{s.t.} \quad ||\Xi'||_p \le \xi 
\vspace{-3pt}
\end{equation} 
where $\Xi$ is updated by $\Xi' \leftarrow P_{\xi}(\Xi + \Delta\Xi)$. 
\begin{figure}[t!]
    \centering
    \includegraphics[width=7.5cm,height=4.2cm]{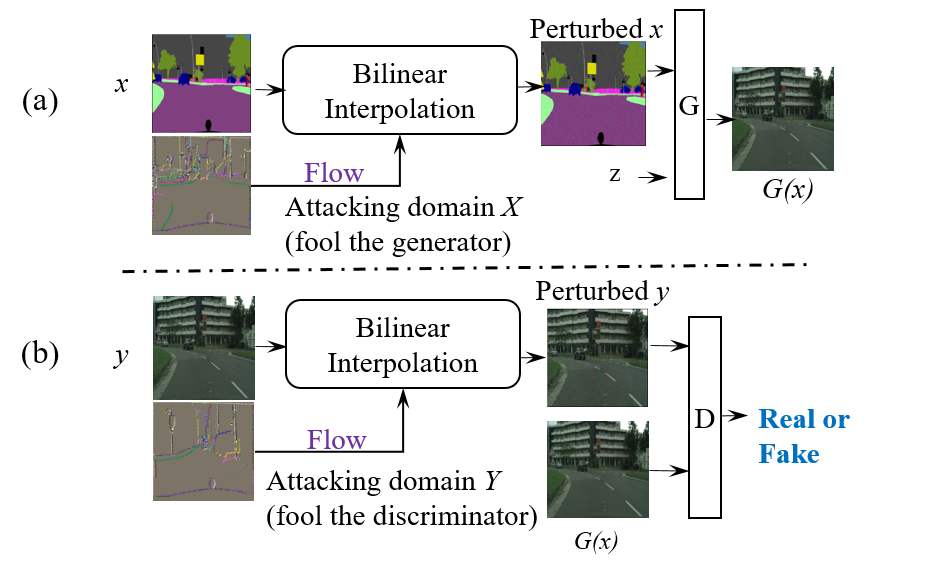}
    \vspace{-5pt}
    \caption{An illustration of flow-based perturbation for fooling Im2Im (paired setting) via both domain $X$ (a) in inference time and $Y$ (b) in training time.\textit{ See more details in the contexts}.}
    \label{fig:spatial_pert}
    \vspace{-15pt}
\end{figure}
Besides, the perturbation we aim to find is also controlled by $\delta$ related to the desired fooling rate for the images sampled from $\mu$ \cite{moosavi2016deepfool}, and satisfies the constraint as follows:
\vspace{-3pt}
\begin{equation}
P_{x\sim \mu} (C(x+\Xi) \ne C(x))\ge 1-\delta 
\vspace{-3pt}
\label{univ_constraint_loss}
\end{equation} 
where $\delta$ quantifies the desired fooling rate for all images sampled from the distribution $\mu$. Note that Eq.~\ref{univ_constraint_loss} is directly optimizable (see Table.~\ref{table:summay_methods}). We then use these misclassified images ($X$ or $Y$) to attack the Im2Im networks. To guarantee that the generated outputs $G(I_{\Xi})$ are mostly mis-classified by the discriminator (see Fig.~\ref{fig:paired_univ_pert} (b)),
we define the objective, which is also optimizable for the input domain:
\vspace{-3pt}
\begin{equation}
\argmin_{\Xi}||\Xi||_p \quad \text{s.t.} \quad D(G(I_{\Xi}))\ne D(G(I)).
\vspace{-3pt}
\label{eq:input_ne}
\end{equation} 
And, for the target domain, the objective is defined as:
\vspace{-3pt}
\begin{equation}
\argmin_{\Xi}||\Xi||_p \quad \text{s.t.} \quad D(I_{\Xi})\ne D(I).
\vspace{-3pt}
\label{eq:target_ne}
\end{equation} 
Hence, when the magnitude of the image-agnostic perturbation is smaller than threshold $\xi$, it is desirable to maintain the high fooling rate for both $X$ and $Y$, which can further fool the discriminator to  wrongly distinguish fake from real and thus allow generating less reliable or wrong outputs. 

\subsection{Flow-based adversarial perturbation} 
The adversarial perturbation can happen not only by pixel value change but also by pixel position shift in an image. Shifting pixel positions will lead to large \emph{$l_p$} norm difference, however, it does not severely affect human perception \cite{xiao2018spatially}. 
Here, we propose a novel flow-based adversarial perturbation aiming to fool Im2Im.

\textit{Our aim is to ultimately fool the generator via flow-based adversarial perturbation such that it fails to generate the expected results.} As mentioned in Sec.~\ref{background_im2im}, we first apply the flow-based adversarial perturbation to the pretrained generator as done in most previous works. We also aim to fool the latent classifier by applying flow-based perturbations to images in the target domain $Y$ in training time, and then feed pristine images in the input domain $X$ to the trained generator. 
We use \emph{$I_{\Xi}^k$} to denote the pixel values of the $k$-th pixel and 2D coordinate (\emph{$u_{\Xi}^k$}, \emph{$v_{\Xi}^k$}) to denote its location in the adversarial example \emph{$I_{\Xi}$}. Here, we assume \emph{$I_{\Xi}^k$} is transformed from the pixel $I^k$ in the original image $I$.  We consider per-pixel flow (displacement) field \emph{f} to synthesize the perturbed image \emph{$I_{\Xi}$} using pixels from the input $I$ in order to fool the generator in Im2Im. 
For the $k$-th pixel within \emph{$I_{\Xi}^k$} at pixel location (\emph{$u_{\Xi}^k$},\emph{$v_{\Xi}^k$}),  we aim to generate perturbations while minimizing the amount of displacement in each image dimension, with the pair denoted by the flow vector $\emph{$f_k$} \mathrel{\mathop:}= (\emph{$\Delta{u^k}$},\emph{$\Delta{v^k}$})$. The flow vector from $f_k$ goes from a pixel in the adversarial example $I_{\Xi}^k$ to its corresponding pixel $I^{k}$ in the input image $I$. As shown in Fig.~\ref{fig:spatial_pert} for paired setting, the location of its corresponding pixel $I^{k}$ can be derived as the $I_{\Xi}$ of the adversarial example as the pixel location of input $I$  can be modeled as 
 (\emph{$u^k$}, \emph{$v^k$}) = (\emph{$u_{\Xi}^k$} + $\Delta{u^k}$, \emph{$v_{\Xi}^k$} + $\Delta{v^k}$) where (\emph{$u^k$}, \emph{$v^k$}) represents the pixel location in the original image. The perturbed image $I_{\Xi}$ is then applied to conditional inputs fed to the generator or directly applied to targets. Note that $k$-th pixel coordinate: ($u^k$, $v^k$) can be fractional numbers and may not lie on the integer image grid. To better optimize the adversarial example \emph{$I_{\Xi}$}, we apply differentiable bilinear interpolation \cite{xiao2018spatially, jaderberg2015spatial} to transform $I$ with flow field $f$. Thus, for the $k$-th pixel and its neighboring pixel $j$, the perturbed image can be formulated as follows:
 \begin{equation}
 \vspace{-3pt}
 \emph{$\emph{$I_{\Xi}$}^k$} = \sum_{j\in N(\emph{$u^k$}, \emph{$v^k$})}\emph{$I^j$}(1-|u^k-u^j|)(1-|v^k-v^j|)
\vspace{-1pt}
\label{flow_adv}
\end{equation} 
where $N({u^k}, v^k)$ are the indices of the 4-pixel neighbors at the location (\emph{$u^k$}, \emph{$v^k$}), namely top-left, top-right, bottom-left and bottom-right. Note that $(1-|u^k-u^j|)(1-|v^k-v^j|)$ is the bilinear sampling kernel defining the gradients with respect to original image $I$ in spatial location.  Thus, we can easily get the flow-based adversarial example $I_{\Xi}$ by calculating the Eq.~\ref{flow_adv}, for every pixel  $I_{\Xi}^k$. 
The goal of spatial flow transformation is to guarantee effective $I_{\Xi}$ while ensuring the magnitude of the perturbation is bounded by a threshold $\xi_{f}$ as in  \cite{carlini2017towards}. The objective for the attack can be formulated the same as the Eq.~\ref{eq:input_ne} and Eq.~\ref{eq:target_ne} while satisfying :
\vspace{-3pt}
\begin{equation}
\arg \min_{f_n} ||f_k- f_n||_2  \quad \text{s.t.} \quad  ||f_n||_p  \le \xi_{f}
\vspace{-3pt}
\label{flow_defn}
\end{equation} 
where $f_n$ is the updated flow from Eq.~\ref{flow_defn}, and the flow loss \emph{$L_{flow}$} can be computed via the sum of spatial displacements. Given a pixel $i$ and neighboring pixels $j\in \emph{N(i)}$, the loss of a flow-based perturbation, which is directly optimizable (see Table.~\ref{table:summay_methods}), is calculated based on the total variation \cite{xiao2018spatially} as:
\vspace{-3pt}
\begin{equation}
\emph{$L_{flow}$} =\sum_{i}^{all}\sum_{j\in \emph{N(i)}} \sqrt{||\Delta{u^i} -\Delta{u^j}||_2^2 + ||\Delta{v^i} -\Delta{v^j}||_2^2}
\vspace{-3pt}
\end{equation}
Thus, the goal of the flow-based perturbation is to fool Im2Im while maintaining the perceptual quality of both the input domain $X$ ( Fig.~\ref{fig:spatial_pert} (a)) and the target domain $Y$ (Fig.~\ref{fig:spatial_pert} (b)).

\subsection{Quasi-physical adversarial perturbation}
\begin{figure}[t]
\centering
    \includegraphics[width=\columnwidth]{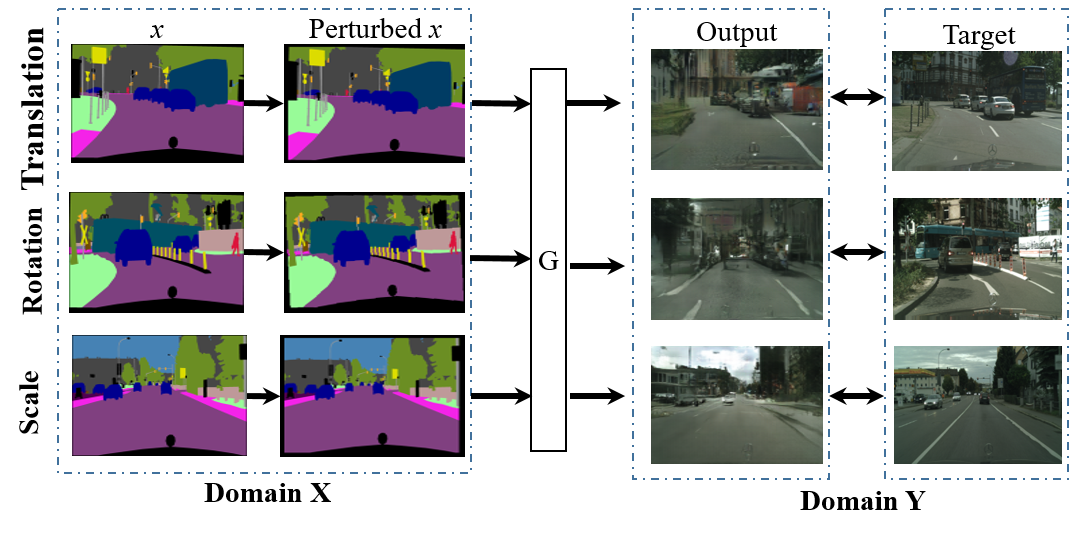}
    \vspace{-15pt}
    \caption{An illustration of quasi-physical adversarial perturbation for Im2Im. Geometrically transformed images in domain A can be adversarial examples for mapping to domain $Y$. When the original segmentation labels (first column) from input domain $X$ are intentionally transformed (translation, rotation and scale) to the adversarial examples (second column), they fail to be mapped to the correct color images in target domain $Y$. }
    \label{fig:rotationImg}
    \vspace{-5pt}
\end{figure}

A physical perturbation may happen when the physical relation between the camera and the scene (\textit{i.e.} distance, viewing position and angle) changes. 
Based on this observation, we propose a quasi-physical adversarial perturbation, which mimics the actual physical perturbation. \textit{We call quai-physical adversarial perturbation since the images are transformed without real change of camera poses but with the scene changes in the digital domain. }
Actually, image transformations such as rotation, translation, and scaling appear natural to human, but we confirm that they alone can cause significant performance deterioration of the Im2Im models as shown in Fig.~\ref{fig:rotationImg}. In this method, \emph{our aim is to fool the  generative model by applying quasi-physical adversarial perturbation}.  A perturbation or an adversarial example can be defined as follows: for a given image $\emph{I}$, when added a perturbation $\Xi$, the image will become an adversarial example \emph{$I_{\Xi}$}, which is the summation as $\emph{I}$ + $\Xi$. In classification problems, adversarial examples are aimed to cause a classifier to misclassify the perturbed image without causing severe perceptible artifacts. Previous works tackle the problem while presuming that the perturbed image \emph{$I_{\Xi}$} and the original image $\emph{I}$ are close if and only if $||$\emph{$I_{\Xi}$}$-\emph{I}||_p$ $\le \xi$ for $\emph{p}$ $\in$ $\{0,\infty\}$ while $\xi$ is small enough. 
However, for the Im2Im problem, we aim to generate quasi-physical perturbations such that undesirable instances are generated as outputs but without causing too much discriminator loss. 
We apply rotation, translation, and scaling, which change given images significantly in terms of $l_p$ norm, however, do not affect human perception much 
as shown in Fig.~\ref{fig:rotationImg}. 
To represent the quasi-physical adversarial attacks to the Im2Im networks in terms of  translation, rotation, and scaling, we define parameters ($\delta\emph{u}$, $\delta\emph{v}$, $\theta$, $s_x$, $s_y$), respectively. 
This similarity transformation can be expressed as follows when the center of image is assumed to be the origin: 
\begin{gather}
\begin{bmatrix} \emph{u'} \\ \emph{v'} \end{bmatrix} =
\begin{bmatrix}
   \emph{$s_x$} &
    0 \\
    0 &
    \emph{$s_y$}
   \end{bmatrix}
   \begin{bmatrix}
   \cos{\theta} &
   -\sin{\theta}\\
   \sin{\theta} &
   \cos{\theta}
   \end{bmatrix} 
   \begin{bmatrix} \emph{u} \\ \emph{v} \end{bmatrix} +
   \begin{bmatrix} \Delta{u}\\ \Delta{v}
   \end{bmatrix}
\end{gather}

By defining the spatial transformation for the attack domain, we construct an adversarial perturbation for $\emph{I}$ by solving the problem for all transformation \emph{$T_S$(
$\delta{u}$, $\delta{v}$, $\theta$, $s_x$, $s_y$)}:\\
\vspace{-4pt}
\begin{equation}
\argmax_{T_S} L(I_{rec},I_{target}) 
\vspace{-8pt}
\end{equation} \\
where $L$ is the loss function, which can be directly obtained from $l_1$ or $l_2$ norm, $I_{rec}$ is the generated image, and $I_{target}$ is the target image to which we want to map. Our aim is to find the minimal spatial transformation that could cause the maximum loss for Im2Im tasks. However, since the loss function is differentiable with respect to inputs but not differentiable with respect to the parameters of transformations, the objective here is not directly optimizable as shown in Table.~\ref{table:summay_methods}.

\begin{table}[t!]
\caption{A summary of the proposed methods regarding their attack domains and whether they are directly optimizable or not. \Checkmark/\xmark indicates yes/no. Method A and B are applied to both domains, and method C is only applied to $X$ and not directly optimizable.}
\vspace{-10pt}
\footnotesize
\begin{center}
\begin{tabular}{c|c|c|c}
\hline
Method & domain X & domain Y & Optimizable \\
\hline\hline
A (image-agnostic)  &\Checkmark & \Checkmark & \Checkmark \\ \hline
B (flow-based)  &\Checkmark & \Checkmark & \Checkmark \\ \hline
C (quasi-physical)  & \Checkmark  & \xmark  & \xmark  \\ 

\hline
\end{tabular}
\end{center}
\vspace{-12pt}
\label{table:summay_methods}
\end{table}

\section{Experiment and evaluation}
We first conduct systematic experiments to substantiate the proposed adversarial perturbations against paired Im2Im network \cite{isola2017image}, using the Cityscapes dataset \cite{cordts2016cityscapes}.
We then conduct experiments using the proposed three methods against the unpaired framework \cite{zhu2017unpaired}, using both Cityscapes and ImageNet datasets \cite{cordts2016cityscapes, krizhevsky2012imagenet}. 
Finally, we evaluate both perceptual realism and performance in real applications (\textit{e.g.} semantic segmentation).
We focus on the analysis of perceptual realism and diversity using the Learned Perceptual Image Patch Similarity (LPIPS)\footnote{LPIPS metric calculates the distance in the AlexNet feature space to better match human perception.} metric proposed in \cite{zhang2018unreasonable}.
To measure whether the generated images with perturbations are able to be used for real tasks or not, we then refer to the FCN-8s for pixel-wise semantic segmentation as in \cite{isola2017image}.

\subsection{ Perturbations under paired setting}

\begin{figure}[t]
\begin{center}
\renewcommand{\tabcolsep}{1pt}
\begin{tabular}{@{}cccc@{}}
   
    \includegraphics[width=20.5mm,height=12.5mm]{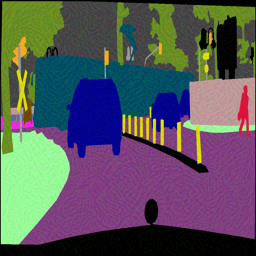}& 
    \includegraphics[width=20.5mm,height=12.5mm]{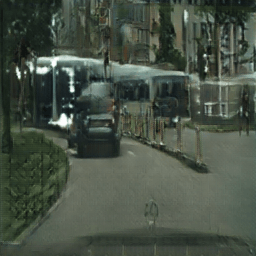} & 
    \includegraphics[width=20.5mm,height=12.5mm]{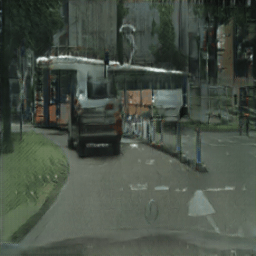} & 
    \includegraphics[width=20.5mm,height=12.5mm]{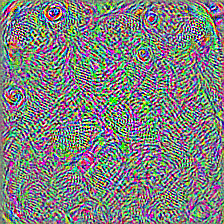}  \\
    \includegraphics[width=20.5mm,height=12.5mm]{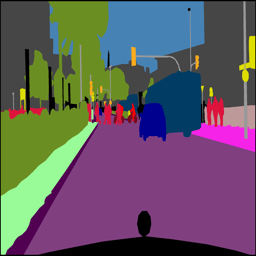}&
    \includegraphics[width=20.5mm,height=12.5mm]{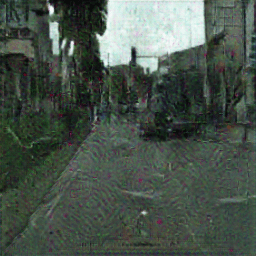}&
    \includegraphics[width=20.5mm,height=12.5mm]{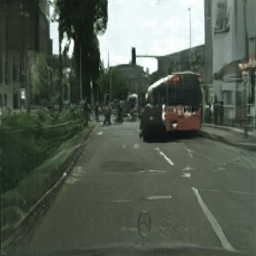}&
    \includegraphics[width=20.5mm,height=12.5mm]{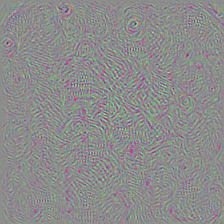}\\
\end{tabular}
\vspace{-5pt}
\caption{Visual results of image-agnostic perturbations on Cityscapes dataset under paired setting, compared with original outputs. The first row shows output when applying perturbations to the input (labels $x$) with \emph{$P_u$}:~$\{\delta,\xi,p \}$=$\{0.2,10,\infty\}$, and the last row depicts the output when applying perturbations to the target images ($y$) with \emph{$P_u$}:~$\{\delta,\xi,p \}$=$\{0.2,1000,\infty\}$.  
From left to right, the perturbed input (first row) and the original input (second row), outputs with perturbation, original outputs, and the visualization of perturbations. }
\label{fig:paired_univ_pert_comp}
\end{center}
\vspace{-10pt}
\end{figure}

\noindent\textbf{Image-agnostic adversarial perturbation.}
In this experiment, we apply image-agnostic adversarial perturbation under paired setting. Results are obtained from $p= \{2,\infty\}$, together with the  $\xi=$ \{10, 200, 500, 1000, 2000\} controlling the magnitude of the perturbation. Meanwhile, we also consider the fooling rate hyper-parameter $\delta =$ \{0.2, 0.6\}, controlling the rate for prior misclassification. Fig.~\ref{fig:paired_univ_pert_comp} and Fig.~\ref{fig:impressive} (second row) show the generated outputs when applying the perturbations to inputs (labels), compared with the original outputs (third column).  
The results show that image-agnostic perturbation is effective. For instance, the unseen perturbations in the inputs and targets cause the erroneous mapping of pedestrians and vehicles, and distortion of roads, etc. The results can also be validated numerically as in Table~\ref{table:paired_universal_compare},  where we can see that most outputs with perturbation ($PO$) are perceptually quite different compared to the original outputs ($O$) as shown in the second column where most LPIPS scores $>0.5$ (the higher LPIPS scores,  the worse and unnatural perceptual realism). Besides, the $PO$s are also quite different from the targets ($I_t$) in domain $Y$. More convincingly, all $PO$s get scores much greater than the reference {0.32 $\pm${0.06}} ($O$ vs $I_t$), which indicates that the outputs are not trustable and unnatural. However, the difference of perturbed images (either inputs or targets) versus the original images  ($PI$ vs $I_{orig}$) is mostly imperceptible to humans (LPIPS $<0.3$).

\begin{table}[t]
\caption{LPIPS scores of image-agnostic perturbations for both domain $X$ (labels $x$) and $Y$ (color images $y$). Convincingly, all the outputs with perturbation (PO) get scores much greater than the reference ({\textbf{0.32 $\pm${0.06}}} ($O$ (original outputs) vs $I_t$ (target color images)), which indicates that the outputs are not trustable and unnatural. \textit{See more details in the contexts}.}%
\vspace{-10pt}
\begin{center}
\footnotesize
\setlength{\tabcolsep}{4pt} 
\begin{tabular}{lccccc}
\hline
$P_u\{\delta,\xi,p \}$ & $PO$ vs $O$ & $PO$ vs $I_t$ & $PI$ vs $I_{orig}$ \\
\hline
$\{0.2,10,\infty \}$ (input)  & 0.32$\pm{0.04}$ & 0.41$\pm{0.06}$ &0.14$\pm${0.03}\\
$\{0.2,10,\infty\}$ (target) & 0.43$\pm{0.04}$ & \textbf{0.54}$\pm{0.05}$ & 0.18$\pm${0.06} \\
$\{0.2, 200,\infty\}$ (input) & 0.41$\pm${0.06} &\textbf{ 0.51}$\pm${0.05}& 0.19$\pm${0.04} \\
$\{0.2, 200,\infty\}$ (target) & \textbf{0.51}$\pm${0.03} & \textbf{0.58}$\pm${0.04}& 0.27$\pm${0.07} \\
$\{0.2,500,\infty\}$ (input) & 0.34$\pm${0.04} & 0.42$\pm${0.06}& 0.20$\pm${0.04}   \\
$\{0.2,500,\infty\}$ (target) & \textbf{0.52}$\pm${0.03} & \textbf{0.58}$\pm${0.04}& 0.27$\pm${0.07} \\
$\{0.2,1000,\infty\}$ (input) & 0.33$\pm${0.04} & 0.43$\pm${0.06}& 0.19$\pm${0.04}   \\
$\{0.2,1000,\infty\}$ (target) & \textbf{0.53}$\pm${0.03} & \textbf{0.60}$\pm${0.03} &0.28$\pm${0.08} \\
$\{0.2,2000,2\}$ (input) & 0.33$\pm${0.03} & 0.43$\pm${0.05}& 0.12$\pm${0.03}\\
$\{0.2,2000,2\}$ (target) & 0.41$\pm${0.03} & \textbf{0.50}$\pm${0.04}& 0.16$\pm${0.05} \\
$\{0.2,2000,\infty \}$ (input) & 0.34$\pm${0.04} & 0.45$\pm${0.06}& 0.20$\pm${0.04}\\
$\{0.2,2000,\infty\}$ (target) & \textbf{0.51}$\pm${0.03} &  \textbf{0.57}$\pm${0.03}& 0.29$\pm${0.08} \\
$\{0.6, 1000,2\}$ (input) & 0.34$\pm${0.04} & 0.42$\pm${0.05} & 0.03$\pm${0.01}\\
$\{0.6,1000,2\}$ (target) & 0.33$\pm${0.04} & 0.44$\pm${0.06}& 0.04$\pm${0.01} \\
\hline
\end{tabular}
\end{center}
\vspace{-20pt}
\label{table:paired_universal_compare}
\end{table}

\begin{figure}[t!]
\begin{center}
\renewcommand{\tabcolsep}{1pt}
\begin{tabular}{@{}cccc@{}}
    
    \includegraphics[width=20.5mm,height=12.5mm]{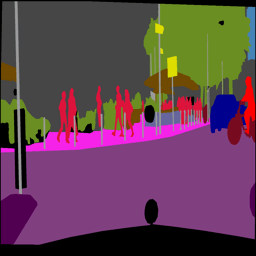}&
    \includegraphics[width=20.5mm,height=12.5mm]{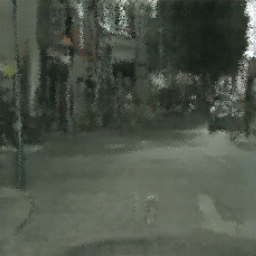}&
    \includegraphics[width=20.5mm,height=12.5mm]{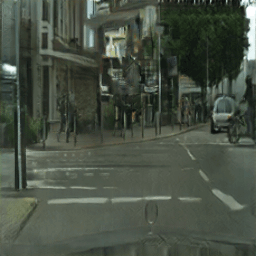}&
    \includegraphics[width=20.5mm,height=12.5mm]{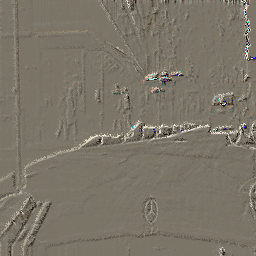} \\
    
    \includegraphics[width=20.5mm,height=12.5mm]{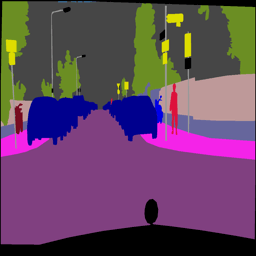}&
    \includegraphics[width=20.5mm,height=12.5mm]{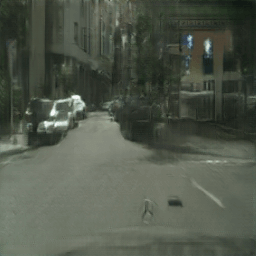}&
    \includegraphics[width=20.5mm,height=12.5mm]{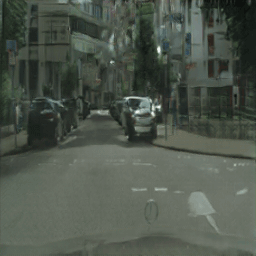}&
    \includegraphics[width=20.5mm,height=12.5mm]{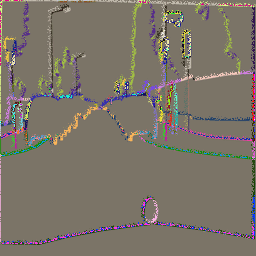}\\
\end{tabular}
\vspace{-5pt}
\caption{Experimental results of flow-based adversarial perturbations on Cityscapes under paired setting. 
From left to right, the perturbed inputs (labels $x$), outputs with perturbations, outputs without perturbations, and visualization of flow, where \emph{ $\xi_{f}$}=$\{1, 2\}$, respectively (from top to bottom). 
Our method is effective to attack the Im2Im and make it yield undesired outputs.}
\label{fig:paired_spatial_pert}
\end{center}
\vspace{-10pt}
\end{figure}

\noindent\textbf{Flow-based adversarial perturbation: }
We then apply flow-based perturbation method on both both inputs (labels $x$) and targets (color images $y$). Fig~\ref{fig:paired_spatial_pert} shows the outputs from our method (second column), compared with the original outputs (third column). We can see that by taking the flow information of pixels, the adversary could cause serious attacks on mappings of vehicles, people, and even lanes from segmentation labels.
Besides, the magnitudes of flows near object boundaries are usually larger than those in other regions, and it indicates that object boundaries are crucial for mapping from one domain to the other.  The results can be also numerically verified from Table~\ref{table:paired_spatial_compare}, where we can see that the outputs with our method ($PO$) are distinctively different from original target color images ($I_t$) and also original outputs (LPIPS$>0.5$), while the perturbed input or target images ($PI$) do not visually differ too much compared to the original images  $I_{orig}$ (LPIPS $<0.25$). Besides, in Table.~\ref{table:paired_spatial_FCN_scores}, outputs with our method yield lower FCN-scores compared to the original outputs ({Per-pixel acc.: 0.66, Per-class acc.: 0.23 and Class IOU:  0.17}).  Meanwhile, perturbations added to labels mostly show higher scores than perturbations to color images, while FCN-scores of both decrease with the increment of flow parameter $\xi_{f}$, and it indicates these color outputs with perturbations severely degrade the performance for the semantic segmentation task.

\begin{table}[t!]
\caption{LPIPS scores of flow-based adversarial perturbations for both domain $X$ and $Y$. The outputs with our methods yield much higher LPIPS scores compared to the original outputs $O$ vs target images $I_t$ (\textbf{0.32$\pm${0.06}})).\textit{ See more details in the contexts}.}
\vspace{-8pt}
\begin{center}
\footnotesize
\begin{tabular}{lcccc}
\hline
${\xi}_f$  & $PO$ vs $I_t$ & $PO$ vs $O$ &  $PI$ vs $I_{orig}$ \\
\hline
1 (input)  & 0.43$\pm{0.05}$ & 0.33$\pm${0.03} &  0.05$\pm{0.01}$ \\
1 (target) & 0.44$\pm{0.05}$ & 0.34$\pm${0.03}  &0.08$\pm{0.02}$ \\
2 (input) & 0.43$\pm${0.05} & 0.33$\pm${0.03} &0.12$\pm{0.02}$ \\
2 (target) & 0.48$\pm${0.06} & 0.40$\pm${0.04} &0.20$\pm{0.04}$ \\
3 (input) & 0.42$\pm${0.05} & 0.33$\pm${0.03}  & 0.19$\pm{0.02}$ \\
3 (target) & 0.53$\pm${0.06} & 0.46$\pm${0.03} & 0.30$\pm{0.05}$ \\
4 (input) & 0.44$\pm${0.06} & 0.34$\pm${0.03}  &  0.24$\pm{0.03}$ \\
4 (target) & \textbf{0.58}$\pm${0.06} & \textbf{0.52}$\pm${0.04}  &  0.39$\pm{0.06}$ \\
\hline
\end{tabular}
\end{center}
\vspace{-10pt}
\label{table:paired_spatial_compare}
\end{table}


\begin{table}[t!]
\footnotesize
\caption{Variation of FCN-scores for the outputs with perturbations added to label (domain $X$) and color (domain $Y$) w.r.t the change of flow of hyper-parameter $\xi_{f}$.  \textit{See more details in the contexts}.}
\vspace{-8pt}
\begin{center}
\begin{tabular}{lcccc}
\hline
${\xi}_f$ & Per-pixel acc. & Per-class acc. &  Class IOU \\
\hline
1 (label)  & 0.37 & 0.18 &  0.12\\
1 (color) & 0.32 & 0.15 & 0.11 \\
2 (label)  & 0.35 & 0.12 & 0.14\\
2 (color) & 0.27 & 0.14 & 0.09\\
3 (label)  & 0.30 & 0.16  & 0.10 \\
3 (color) & 0.23 & 0.11 & 0.05\\
4 (label)  & \textbf{0.30} & \textbf{0.15} & \textbf{0.08}\\
4 (color) & \textbf{0.22} & \textbf{0.10} & \textbf{0.09} \\
Original output  & 0.66 & 0.23 & 0.17\\
\hline
\end{tabular}
\end{center}
\vspace{-25pt}
\label{table:paired_spatial_FCN_scores}
\end{table}

\begin{figure}[t!]
\begin{center}
\renewcommand{\tabcolsep}{1pt}
\begin{tabular}{@{}cccc@{}}
    \includegraphics[width=20.5mm,height=12.5mm]{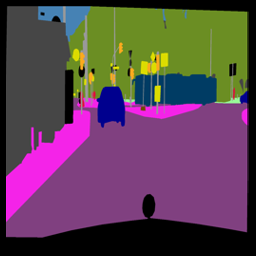}&
    \includegraphics[width=20.5mm,height=12.5mm]{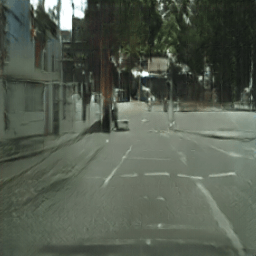}&
    \includegraphics[width=20.5mm,height=12.5mm]{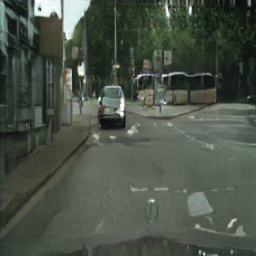}&
    \includegraphics[width=20.5mm,height=12.5mm]{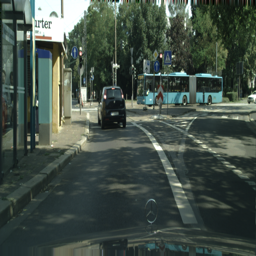}\\
\end{tabular}
\vspace{-5pt}
\caption{Experimental results for the quasi-physical adversarial perturbation. First column shows perturbed inputs ($S$=0.95), followed by output with our method, original output and target color image.  From the output with perturbation, we can see small transformations can cause serious attacks (fail to map people and vehicles).}
\label{fig:paired_physical_pert}
\end{center}
\vspace{-10pt}
\end{figure}

\noindent\textbf{Quasi-physical adversarial perturbation.}  In order to maintain the visual similarity of perturbed images to the original images, we restrict the magnitude of allowed perturbations to be relatively small for both conditional inputs and targets, without applying any modifications of pixel values. We consider $\ang{0.5}$ rotation ($R$) each time, translations ($T$) less than around 10 $\%$ of a network input image size, and scaling ($S$). More explicitly, we experiment on the perturbations with $R=\{0^\circ, 0.5^\circ, 1^\circ, 1.5^\circ, 2^\circ, 2.5^\circ, 3^\circ, 3.5^\circ\}$ combined with $S=\{0.85, 0.95\}$ and $T=\{10, 20\}$ (in pixels). When applying scale, rotation, and translation, the blank regions are all filled with black pixels. 
As shown in Fig.~\ref{fig:paired_physical_pert}, we can see that tiny magnitude of perturbation adding to inputs (labels) could cause serious attacks in the outputs, such as missing building, vehicles, pedestrians, traffic lights, etc. This reflects that current Im2Im networks are vulnerable to physical and quasi-physical adversarial perturbations and need to be improved to deal with such perturbations. Numerically, it also turns out that outputs with our perturbation ($PO$) are quite visually different from the original outputs ($O$) and also the targets $I_t$ (color images), which indicates that the outputs are unnatural. 
\subsection{Perturbations under unpaired setting }
 Unpaired mapping has been applied to generate diverse driving environments via style transfer (\textit{e.g.} day to night, winter to spring) \cite{ zhu2017unpaired,huang2018multimodal} or domain adaptation of semantic segmentation for autonomous vehicles \cite{zou2018unsupervised}. In this experiment, we show the experimental results on the baseline structure \cite{zhu2017unpaired} using the Cityscapes and ImageNet datasets.  Here, we show the robustness of our methods based on both datasets to attack example-guided style transfer, which aims at modifying the style of an image while preserving its content. 
 
\begin{figure}[t!]
\begin{center}
\renewcommand{\tabcolsep}{1pt}
\begin{tabular}{@{}ccccc@{}}
    
    \includegraphics[width=16.2mm,height=12mm]{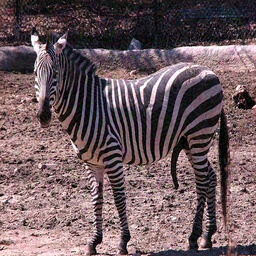}&
    \includegraphics[width=16.2mm,height=12mm]{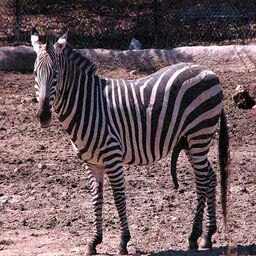}&
    \includegraphics[width=16.2mm,height=12mm]{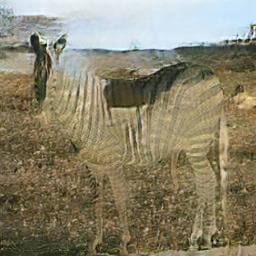}&
    \includegraphics[width=16.2mm,height=12mm]{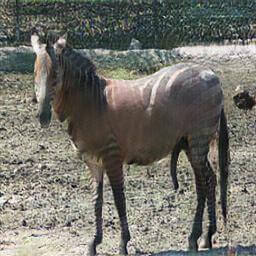}&
    \includegraphics[width=16.2mm,height=12mm]{figures/delta_02xi_10p_inf.png}\\
    \includegraphics[width=16.2mm,height=12mm]{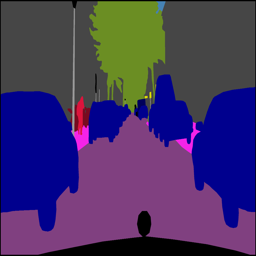}&
    \includegraphics[width=16.2mm,height=12mm]{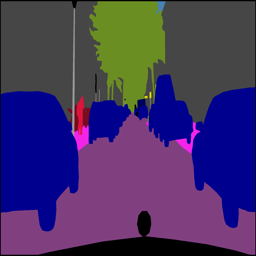}&
    \includegraphics[width=16.2mm,height=12mm]{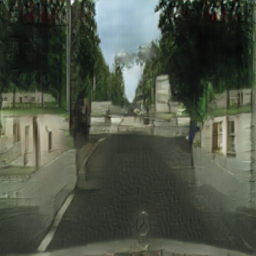}&
    \includegraphics[width=16.2mm,height=12mm]{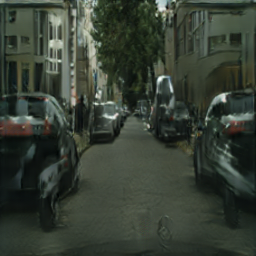}&
    \includegraphics[width=16.2mm,height=12mm]{figures/fig0/delta_02xi_2000p_inf.png}\\
\end{tabular}
\vspace{-5pt}
\caption{Experimental results of image-agnostic adversarial perturbations on Im2Im for style transfer. From left to right, original input, perturbed input, output with our method, original outputs, and visualization of perturbations ($P_u\{\delta,\xi,p \}$ = $\{0.2,10,\inf \}$ and $\{0.2,2000,\inf \}$). \textit{See more details in the contexts}.
}
\label{fig:unpaired_univ_pert}
\end{center}
\vspace{-18pt}
\end{figure}

\noindent\textbf{Image-agnostic adversarial perturbation.}
We set  hyper-parameters as $p= \{2, \infty\}$, together with the  $\xi=$ \{10, 500, 1000, 2000\} and fooling rate hyper-parameter $\delta =$ \{0.2, 0.6\}. 
In Fig.~\ref{fig:unpaired_univ_pert}, we can clearly see when the perturbation is applied, a zebra fails to be transferred to horse. This indicates our method is effective enough to impede the style transfer in autonomous driving (\textit{e.g.} fail to generate diverse driving environments). In the second row, we can see  that there exists image-agnostic perturbation which could demolish the transfer from input (label) to the target (color image) as shown in the third column, compared with the original output in the fourth column. The existence of adversarial perturbation causes serious attack on the mapping of cars, lane, building from input. 

\begin{figure}[t!]
\begin{center}
\renewcommand{\tabcolsep}{1pt}
\begin{tabular}{@{}ccccc@{}}

    \includegraphics[width=16.2mm,height=12mm]{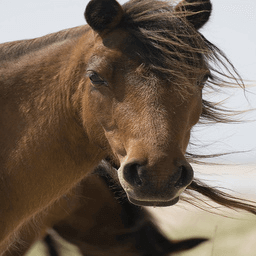}&
    \includegraphics[width=16.2mm,height=12mm]{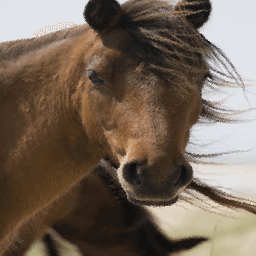}&
    \includegraphics[width=16.2mm,height=12mm]{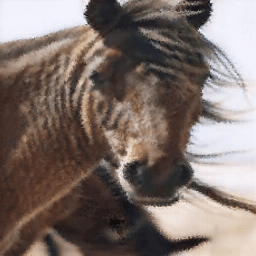}&
    \includegraphics[width=16.2mm,height=12mm]{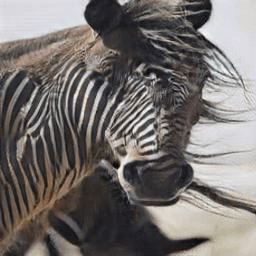}&
    \includegraphics[width=16.2mm,height=12mm]{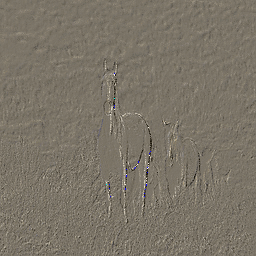}\\
    \includegraphics[width=16.2mm,height=12mm]{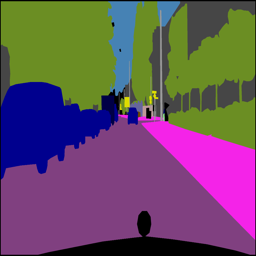}&
    \includegraphics[width=16.2mm,height=12mm]{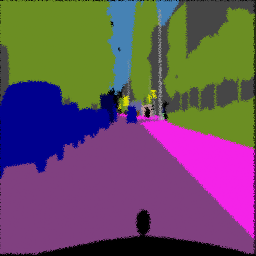}&
    \includegraphics[width=16.2mm,height=12mm]{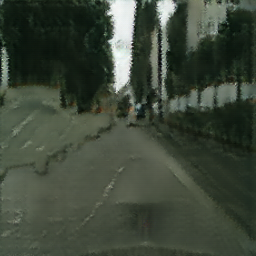}&
    \includegraphics[width=16.2mm,height=12mm]{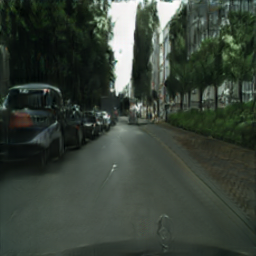}&

    \includegraphics[width=16.2mm,height=12mm]{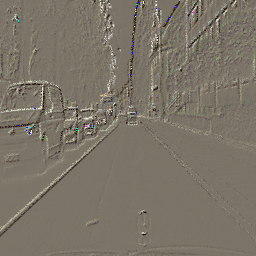}\\
    
   
\end{tabular}
\vspace{-5pt}
\caption{Experimental results of flow-based adversarial perturbations on Im2Im for style transfer. The first column shows the pristine horse image, followed by the flow-transformed images with tiny perturbations shown in the last column ($f_{\xi}=1$ and $f_{\xi}=2$). The third column depicts the outputs with our method, compared with the original outputs in the fourth column.}
\label{fig:unpaired_pert_flow}
\end{center}
\vspace{-10pt}
\end{figure}

\noindent\textbf{Flow-based adversarial perturbation.} 
We have also tested the unpaired mapping with flow-based perturbations on Im2Im style transfer. We assign hyper-parameter \emph{ $\xi_{f}= \{1,2,3,4\}$} to flow-based perturbations for both domains (inputs and targets).
Fig.~\ref{fig:unpaired_pert_flow} and Fig.~\ref{fig:impressive}(third row) show some examples of outputs with flow-based perturbation. 
As can be seen in Fig.~\ref{fig:unpaired_pert_flow}, the outputs with our method are still horse images, not zebra images as we expect. Meanwhile, the input also fails to be mapped to the target (third column). More serious situations happen with larger flow transformations, which would even cause serious attacks on the mapping of patterns. The effectiveness of our method can also be numerically verified in Table.~\ref{table:unpaired_spatial_comp}, where the outputs with perturbation ($PO$) are semantically and textually different from original outputs ($O$) (LPIPS $\ge$ 0.5). However, the perturbed input ($PI$) are visually similar to the original images ($I$) (LPIPS $\le$ 0.25). The results also reflect that flow-based perturbation is effective on fooling Im2Im for autonomous driving. For instance, the adversarial examples might impede the transfer for driving scenes (e.g. night to daytime). Besides, they can also cause serious attacks on the transfer from simulated domain to real domain.   

\begin{table}[t!]
\caption{LPIPS scores of flow-based adversarial perturbations under unpaired setting. See more details in the contexts.}
\vspace{-10pt}
\begin{center}
\footnotesize
\begin{tabular}{p{1.4cm}p{1.5cm}p{1.5cm}p{1.8cm}p{1.5cm}}
\hline
$\xi_{f}$ &  \multicolumn{2}{l}{PO vs O} & PI vs I \\
\hline
 & AtoB & BtoA & Pert. ratio \\
 \hline                                                                           
1 (A) & 0.43 $\pm{0.05}$ & 0.45$\pm${0.07} & 0.09 $\pm{0.05}$\\
1 (B) & 0.41 $\pm{0.05}$ & 0.38$\pm${0.05} & 0.12 $\pm{0.04}$\\
2 (A) & 0.50 $\pm${0.08} & 0.51$\pm${0.07} & 0.18 $\pm{0.05}$\\
2 (B) & 0.49 $\pm${0.07} & 0.47$\pm${0.07} & 0.19 $\pm{0.09}$\\
3 (A) & \textbf{0.52} $\pm${0.07} & \textbf{0.53}$\pm${0.07} & \textbf{0.23} $\pm{0.08}$\\
3 (B) & \textbf{0.51} $\pm${0.05} & \textbf{0.50}$\pm${0.07} & \textbf{0.25} $\pm{0.09}$\\
\hline
\end{tabular}
\end{center}
\vspace{-10pt}
\label{table:unpaired_spatial_comp}
\end{table}
\begin{figure}[t!]
\begin{center}
\renewcommand{\tabcolsep}{1pt}
\begin{tabular}{@{}cccccc@{}}
    \includegraphics[width=13.5mm,height=12mm]{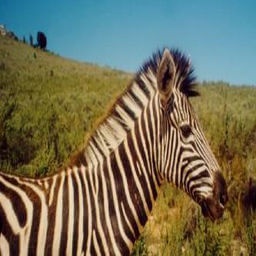}&
    \includegraphics[width=13.5mm,height=12mm]{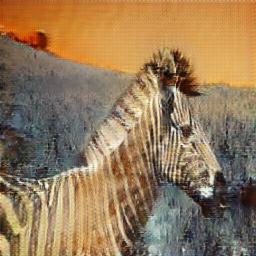}&
    \includegraphics[width=13.5mm,height=12mm]{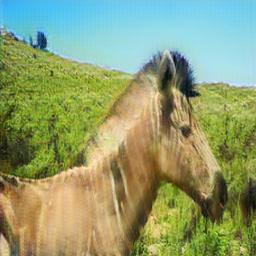}&
    \includegraphics[width=13.5mm,height=12mm]{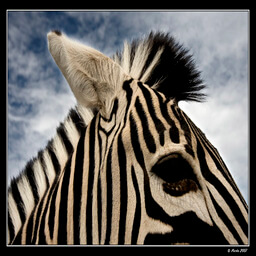}&
    \includegraphics[width=13.5mm,height=12mm]{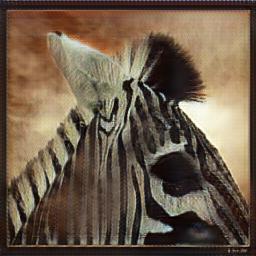}&
    \includegraphics[width=13.5mm,height=12mm]{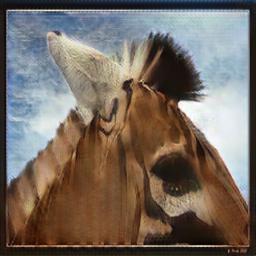} \\
\end{tabular}
\vspace{-5pt}
\caption{Visual results with quasi-physical adversarial perturbations for style transfer on ImageNet dataset. From left to right, the first and fourth columns show the inputs, and the second and fifth columns show the perturbed outputs with rotated horse image ($2^\circ$), compared with original outputs in the third and sixth columns. }
\label{fig:unpaired_physical_pert}
\end{center}
\vspace{-23pt}
\end{figure}

\begin{figure}[t!]
\vspace{-5pt}
    \centering
    \includegraphics[ width=\columnwidth, height=5.8cm]{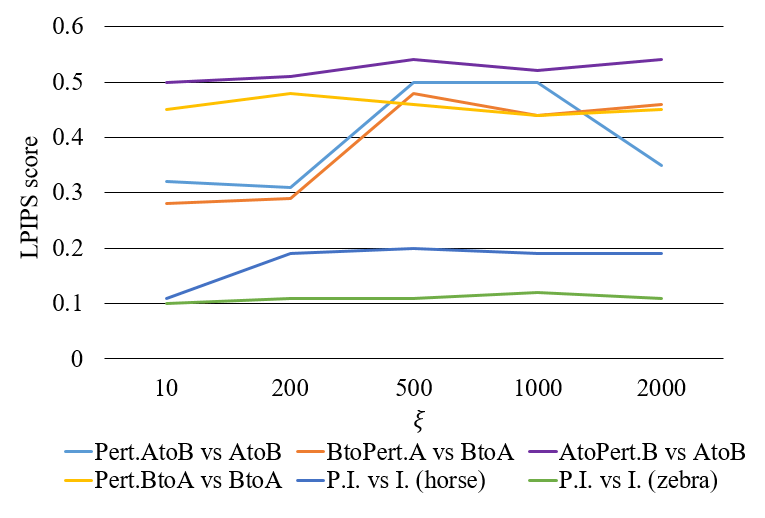}
    \vspace{-20pt}
    \caption{Variation of LPIPS scores with respect to the change of the image-agnostic perturbation hyper-parameter $\xi$ tested on \emph{horse2zebra} examples. Note that Pert.AtoB indicates the mapping output of the perturbed horse images (domain A) to zebras (domain B), and others (\textit{e.g.} Pert.BtoA) have similar indications. Pert.AtoB vs AtoB means the perceptual comparison between the outputs of Pert.AtoB and the outputs of AtoB (horse to zebra), and other notations are of similar indications. \textit{See more details in the contexts.}}
\vspace{-15pt}
\label{fig:LPIPS_score_unpaired_xi}
\end{figure}

\noindent\textbf{Quasi-physical adversarial perturbation.}
We again apply the quasi-physical adversarial perturbation style transfer task. 
Fig.~\ref{fig:unpaired_physical_pert} shows some outputs from quasi-physical adversarial perturbations. 
As can be seen, a simple rotation with $2^\circ$ of horse images (domain B) disrupts the mapping, which makes a zebra still mapped to a zebra. The experiment on Cityscapes dataset under unpaired setting also show the similar results as shown in Fig.~\ref{fig:paired_physical_pert} under paired setting. The results indicate that our method could also cause attacks on soiling and adverse weather classification for autonomous driving as done in \cite{uricar2019yes}.   

\section{Discussion}
\noindent \textbf{Generalization.} We have tested the perturbations on both paired and unpaired settings on various datasets. Our method is effective to attack Im2Im (\textit{e.g.} labels fail to be mapped to color images with loss of vehicles, people and lanes, and style transfer (one pattern to another) for autonomous driving applications. In the future work, it is plausible to apply defense method based on our attack methods. 
 
\noindent\textbf{How adversarial perturbations affect Im2Im?}
We have shown that our proposed methods are effective enough to fool the Im2Im frameworks for autonomous driving. Our methods can illustrate how the perceptual quality of outputs varies with respect to the change of perturbations.
Take image-agnostic adversarial perturbation as an example,
from the LPIPS scores shown in Fig~\ref{fig:LPIPS_score_unpaired_xi},
we can see that the perturbations have the highest fooling rate when $\xi=500$ (threshold), where LPIPS scores of most outputs (horse2zebra (AtoB) and zebra2horse (BtoA)) culminate. This indicates that, when $P_u=\{0.2, 500, \infty \}$, the most unnatural images are generated. Meanwhile, when $\xi$ increases from $500$, the LPIPS score decreases, and it shows that the mapping is rather disrupted. However, the LPIPS scores between perturbed images ($PI$) and original images ($I$) are almost flattened, indicating perturbed images are natural and perturbations are imperceptible. 

\noindent\textbf{Difference from classification problems. } Adversarial attack for DNN aimed for classification is essentially deterministic and objective than Im2Im frameworks (without class labels). Attacking Im2Im is more knotty since it is from one domain to the other, which renders the problem challenging. We thus propose to apply perturbation mechanisms to the discriminator in training phase since the target images are fed to it to tell real from fake. By examining the discriminator towards adversarial perturbations, it help us to understand more clearly about Im2Im networks and their vulnerability. 

\noindent \textbf{New perspectives for Im2Im.}
In this work, the study of adversarial perturbations for Im2Im provides us new outlooks on how to improve the design of Im2Im networks and the preparation of dataset. First of all, as shown in Fig.~\ref{fig:paired_univ_pert}, \ref{fig:paired_spatial_pert}, perturbations to target domain causes more severe degradation of outputs, thus it is better to design more robust discriminator to better distinguish fake from real. Second, since physical perturbation has strong impact on the outputs of Im2Im (Fig.~\ref{fig:paired_physical_pert} and Fig.~\ref{fig:unpaired_physical_pert}),  it is better to avoid using the scenes in images that have large view transformations. Third, since we have showed that failure of outputs is also caused by the perturbations intentionally or unintentionally added to the dataset, it is crucial to apply defensive approaches to the dataset to used for training and test. Lastly, since the generator is sensitive to the perturbations to the input, it is better to design the generator that takes additional information for better domain mapping. 

\section{Conclusion and Future Work} 
In this paper, we have demonstrated how our proposed methods cam successfully fool the Im2Im frameworks for autonomous driving applications. We first proposed three novel and initiative perturbation methods for both paired (labels $\to$ color) and unpaired settings (style transfer). We then conducted intensive experiments to validate our methods and evaluated the results using various perceptual realism metrics. Since we conventionally trust what we see from the outputs, perturbations that can easily demolish mapping from domain to domain and impede the utilization of generated results should be considered as threatening factors for autonomous vehicles. It turns out that these visually imperceptible or quasi-imperceptible perturbations could actually provide potential outlooks for the security of applying Im2Im networks to autonomous driving research. We are currently exploiting to improve the robustness of Im2Im networks for more secure autonomous driving and robotic applicaions.



\ifCLASSOPTIONcaptionsoff
  \newpage
\fi

{\small 
\bibliographystyle{plain}
\bibliography{root}
}

\end{document}